\documentclass{article}
\usepackage[preprint]{neurips_2020}  

\usepackage[utf8]{inputenc}
\usepackage[T1]{fontenc}
\usepackage[hyphens]{url}
\usepackage{graphicx}
\usepackage{booktabs}
\usepackage{array}
\usepackage{amsmath,amssymb}

\title{Compression Technique Interactions in Small Language Models:\\ A Factorial Ablation Study}

\author{%
  Bhushit Nema\thanks{Equal contribution.} \\
  Amazon Web Services \\
  Seattle, WA, USA \\
  \texttt{bnema@amazon.com} \\
  \And
  Anjali Kadiyala\footnotemark[1] \\
  Amazon Web Services \\
  Seattle, WA, USA \\
  \texttt{anjkdyl@amazon.com} \\
  \AND
  Siddharth Krishnamurthy\footnotemark[1] \\
  Amazon Web Services \\
  Seattle, WA, USA \\
  \texttt{siddhakr@amazon.com} \\
}

\begin{document}

\maketitle

\begin{abstract}
Deploying small language models on resource-constrained hardware requires compression, yet practitioners lack systematic guidance on how quantization, pruning, and post-compression recovery interact when combined. We present a factorial ablation study evaluating 72 compression configurations across four SLM architectures (2.7B--7B parameters), extended with a 48-configuration ordering analysis, benchmark accuracy validation on 30 compressed models, and robustness checks including multi-seed validation and full-corpus perplexity evaluation. INT8 quantization is effectively lossless ($<$0.5\% accuracy degradation), while NF4 quantization combined with L1 magnitude pruning causes model collapse in all 16 tested combinations. LoRA-based fine-tuning reduces degraded perplexity by up to 97\% after aggressive pruning; an ablation against standard fine-tuning without a teacher model shows the recovery is driven by LoRA adaptation itself rather than the distillation signal. Compression ordering shows no significant effect among viable configurations (ANOVA $p = 0.96$) but determines viability for NF4+pruning combinations. INT8 quantization alone provides 40--45\% memory reduction with no quality loss. On-device CPU deployment via llama.cpp confirms these findings transfer off the GPU: 4-bit quantization runs 46--54\% faster than 8-bit for a perplexity difference under 0.5 points, while unstructured pruning yields no CPU latency or memory benefit, consistent with its dense storage format.
\end{abstract}

\section{Introduction}

In practice, when building and deploying language models across cloud, on-premise, and edge environments, the question is rarely \textit{whether} to compress a model but rather which combination of techniques to apply and in what order. A 7B parameter model in half-precision occupies roughly
14GB of GPU memory, which exceeds the capacity of most embedded GPUs, mobile devices, and even some cloud inference endpoints. As a result, compression is not optional for most production deployments of small language models in the 2--7B parameter range.

Three compression techniques dominate the practitioner's toolkit: quantization (reducing weight precision from 16-bit to 8-bit or 4-bit), pruning (zeroing out low-magnitude weights), and knowledge distillation (training a compressed model to match a teacher's output distribution, commonly applied to recover quality lost to compression). Each is well-studied in isolation~\citep{dettmers2022llmint8,frantar2023sparsegpt,hinton2015distilling}. What is missing is systematic evidence on how they behave in combination. Does quantizing a pruned model produce the same result as pruning a quantized model? Can distillation recover quality lost from both simultaneously? At what point does stacking cause outright failure? Prior work evaluates these techniques individually or in fixed pairs, but real deployments rarely use just one: a team targeting a Jetson Nano with 4GB of shared memory will likely need both quantization and pruning to fit a 4B model, and may want distillation to recover whatever quality was lost. A team deploying on an edge server with 16GB may only need quantization but wants to know whether adding pruning helps or hurts. Without systematic data, these decisions are made through trial and error.

We address this gap with a controlled factorial study. The ablation spans 72 compression configurations across four model architectures, with an additional 48 runs testing whether the sequence of compression operations matters. All experiments run on a single NVIDIA L40S GPU (ml.g6e.2xlarge cloud instance), ensuring consistent and comparable measurements. We validate our primary metric (WikiText-2 perplexity) against benchmark accuracy (MMLU, HellaSwag, ARC, BoolQ) on 30 compressed configurations, and confirm reproducibility through multi-seed validation across three random seeds. The study produces six principal contributions: (1)~a complete $3 \times 3 \times 2$ factorial design across four architectures, (2)~a 48-configuration ordering analysis, (3)~LoRA-based fine-tuning as a post-compression recovery mechanism with an ablation showing the teacher signal is not the critical component, (4)~benchmark accuracy validation confirming perplexity as a reliable proxy, (5)~on-device CPU deployment measurements (llama.cpp) across four memory tiers, and (6)~robustness validation demonstrating seed independence and full-corpus perplexity consistency ($r = 0.985$ between 20-sample and 1{,}940-sample evaluation).
\section{Related Work}

\subsection{Quantization and Pruning}

Post-training quantization methods---GPTQ~\citep{frantar2022gptq}, AWQ~\citep{lin2023awq}, SmoothQuant~\citep{xiao2023smoothquant}, and bitsandbytes~\citep{dettmers2022llmint8,dettmers2023qlora}---each report modest accuracy loss at INT8 or INT4 precision when applied individually, but none examine interactions with pruning or distillation. On the pruning side, SparseGPT~\citep{frantar2023sparsegpt} and Wanda~\citep{sun2023wanda} achieve 50--60\% sparsity with minimal degradation, while LLM Pruner~\citep{ma2023llmpruner} uses gradient information to identify structurally coupled parameter groups. These works pair pruning with retraining but do not include quantization as a simultaneous factor. Structured approaches such as head pruning require architecture-specific module knowledge that varies across model families.

\subsection{Distillation and Combined Compression}

\citet{hinton2015distilling} established knowledge distillation; DistilBERT~\citep{sanh2019distilbert} applied it to produce a compact BERT variant. QLoRA~\citep{dettmers2023qlora} enables fine-tuning of quantized models through LoRA~\citep{hu2021lora}, and KD-LoRA~\citep{rambhatla2024kdlora} combines LoRA with a distillation objective. Born-Again Networks~\citep{furlanello2018born} showed that self-distillation can match or surpass teacher accuracy; our work applies self-distillation as a post-compression repair step. ZipLM~\citep{kurtic2023ziplm} and SqueezeLLM~\citep{kim2023squeezellm} combine quantization with pruning but do not systematically vary all three techniques or test ordering.

Three recent works are closest to ours. Minitron~\citep{muralidharan2024minitron} combines width and depth pruning with distillation-based retraining, producing compressed models that outperform same-sized models trained from scratch, but omits quantization. UniComp~\citep{vonrad2026unicomp} evaluates pruning, quantization, and distillation across 40 benchmarks but compares techniques
individually rather than studying their interactions when stacked. \citet{chhawri2025ordering} study compression ordering on a single 3B model with structured pruning, finding that pruning before quantization preserves quality. Our study differs in scope: we test four architectures with unstructured pruning in a full factorial design, identify sharp viability
boundaries (NF4+pruning collapse), and include a LoRA recovery ablation that disentangles the teacher signal from the adaptation effect.
\section{Experimental Setup}

\subsection{Models}

We selected four models spanning three providers, three architecture families, and a 2.6$\times$ parameter range (Table~\ref{tab:models}). Including two models at the 4B scale from different providers tests whether compression behavior is architecture-dependent at the same parameter count.

\begin{table}[h]
  \caption{Target models. All loaded from Hugging Face Hub.}
  \label{tab:models}
  \centering
  \small
  \begin{tabular}{llll}
    \toprule
    Model & Provider & Params & License \\
    \midrule
    Phi-2 & Microsoft & 2.7B & MIT \\
    Qwen3-4B & Alibaba & 4.0B & Apache-2.0 \\
    Nemotron-Mini-4B & NVIDIA & 4.0B & NVIDIA OML \\
    Qwen2.5-7B-Inst. & Alibaba & 7.0B & Apache-2.0 \\
    \bottomrule
  \end{tabular}
\end{table}

\subsection{Compression Techniques}

\paragraph{Quantization.} We use bitsandbytes~\citep{dettmers2022llmint8} for INT8 and INT4 (NF4) quantization. The NF4 format minimizes quantization error for approximately Gaussian weight distributions~\citep{dettmers2023qlora}.

\paragraph{Pruning.} We apply unstructured L1 magnitude pruning at 30\% and 50\% sparsity using PyTorch's pruning utilities. Structured pruning (attention head removal) was inapplicable to three of four architectures due to fused attention projections, and destructive on the fourth (Appendix). Unstructured pruning does not directly translate to inference speedup without sparse computation kernels.

\paragraph{Post-Compression Recovery.} We add LoRA adapters~\citep{hu2021lora} ($r = 8$, $\alpha = 16$, targeting q\_proj and v\_proj), trained for 200 steps at learning rate $2 \times 10^{-4}$ on 256 C4 samples~\citep{raffel2020t5} truncated to 128 tokens. The primary study uses self-distillation (KL divergence against the model's own FP16 logits, $T = 2$, $\lambda = 0.5$); a separate ablation compares this against LoRA-only fine-tuning (cross-entropy, no teacher) to isolate the teacher signal's contribution (Section~\ref{sec:distill_ablation}). We use self-distillation rather than cross-architecture distillation to isolate the recovery effect: any improvement comes from recovering compressed-away knowledge, not from transferring new knowledge from a more capable teacher, keeping the recovery factor in our factorial design clean to interpret. The LoRA approach is necessary because quantized weights are frozen; the adapter matrices (${\sim}$0.1\% of total parameters) provide a trainable pathway without modifying the quantized base. LoRA configuration is fixed to common defaults~\citep{hu2021lora,dettmers2023qlora} across all cells.

\subsection{Factorial Design}

The primary study uses a $3 \times 3 \times 2$ factorial design: three quantization levels (none, INT8, INT4), three pruning levels (none, 30\%, 50\%), and two recovery conditions (off, on), producing 18 configurations per model and 72 total with fixed ordering $Q \rightarrow P \rightarrow R$. The ordering study tests three alternative orderings on the 16 configurations where all three factors are active, across all four models, producing 48 additional runs. Total: 120 configurations.

\subsection{Evaluation}

WikiText-2 perplexity~\citep{merity2016pointer} serves as the primary metric, computed on 20 samples with 512-token context (${\sim}$30 seconds per model versus 30--60 minutes for benchmark suites). We validate the subset against the full test set (1{,}940 samples; $r = 0.985$) and against benchmark accuracy (MMLU~\citep{hendrycks2020mmlu}, HellaSwag~\citep{zellers2019hellaswag}, ARC-Challenge~\citep{clark2018arc}, BoolQ~\citep{clark2019boolq}) on 30 configurations via lm-eval~\citep{gao2023lmeval}. We also measure GPU memory and latency (mean of 5 runs, 50 tokens, batch size 1). The primary study uses seed 42; multi-seed validation on 16 key configurations across three seeds confirms stability ($\sigma \leq 0.7$ PPL points; Appendix).

\subsection{Infrastructure}

All experiments run on a single NVIDIA L40S GPU (48GB VRAM, ml.g6e.2xlarge). The L40S accommodates all four models in FP16 with headroom for self-distillation (Qwen2.5-7B student plus teacher totals 28GB).

\subsection{On-Device CPU Deployment}
\label{sec:edge_setup}

To validate GPU findings on real hardware, we convert each model to GGUF format and run under llama.cpp with q8 and q4 quantization on CPU. We measure peak RSS, generation throughput (mean of five runs), and perplexity, recording whether each model fits within four memory tiers (2, 4, 8, 12 GB). CPU perplexity is not directly comparable to GPU perplexity due to different tokenization; we report only within-engine comparisons. Model rankings are preserved across engines (Spearman $\rho = 1.0$).

\section{Results}

\subsection{Baseline Performance}

Table~\ref{tab:baseline} shows uncompressed model performance. Qwen2.5-7B-Instruct leads on both MMLU (0.718) and perplexity (11.82). The two 4B models show notably different accuracy profiles despite having the same parameter count: Qwen3-4B achieves substantially higher MMLU (0.684 vs 0.569) while Nemotron-Mini-4B has better perplexity (18.53 vs 26.72).

\begin{table}[h]
  \caption{Baseline performance (uncompressed FP16). Memory in GB.}
  \label{tab:baseline}
  \centering
  \small
  \begin{tabular}{lcccccc}
    \toprule
    Model & PPL & MMLU & HSwag & ARC & BoolQ & Mem \\
    \midrule
    Phi-2 & 20.7 & .545 & .559 & .530 & .834 & 5.2 \\
    Qwen3-4B & 26.7 & .684 & .523 & .508 & .851 & 7.5 \\
    Nemotron & 18.5 & .569 & .559 & .506 & .826 & 7.8 \\
    Qwen-7B & 11.8 & .718 & .621 & .527 & .864 & 14.2 \\
    \bottomrule
  \end{tabular}
\end{table}

\subsection{Quantization and Pruning Interactions}

Of 72 configurations, 55 produced viable models (perplexity below 1000) and 17 resulted in model collapse. All 16 NF4+pruning configurations collapsed (4 models $\times$ 2 pruning levels $\times$ 2 recovery states). One additional collapse occurred: Nemotron-Mini-4B at 50\% pruning without recovery (perplexity 1{,}108). No INT8 configuration collapsed at any pruning level.

INT8 quantization alone barely affects perplexity, staying within 1 point of the FP16 baseline across all models. INT4 without pruning is similarly well-behaved (1--2 point increases). Pruning at 30\% causes moderate degradation (perplexity 22--28). At 50\% sparsity, FP16 models show severe degradation (perplexity 56--1108), while INT8 models tolerate it surprisingly well (perplexity 22--31).

\begin{figure}[h]
  \centering
  \includegraphics[width=\columnwidth]{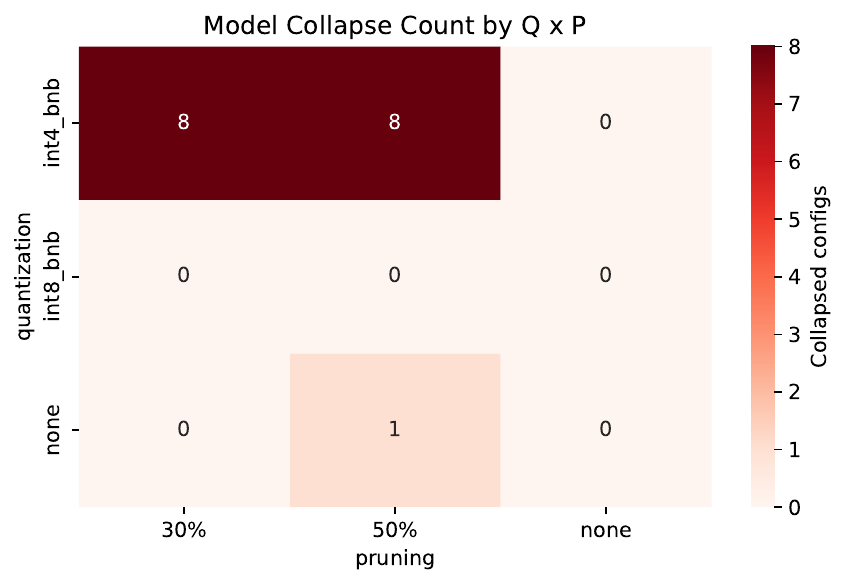}
  \caption{Model collapse count by quantization and pruning level. All 16 NF4+pruning configurations collapsed. No INT8 configuration collapsed.}
  \label{fig:collapse}
\end{figure}

The critical finding is the NF4 + pruning incompatibility. NF4 weights use only 4 bits per parameter; removing 30--50\% of these already heavily compressed weights leaves insufficient information for coherent generation. INT8 weights, with twice the precision, contain enough redundancy to survive the same pruning levels. Figure~\ref{fig:collapse} visualizes this pattern.

\subsection{Post-Compression Recovery via LoRA}

LoRA-based fine-tuning is remarkably effective at recovering models after aggressive pruning. Table~\ref{tab:recovery} shows recovery rates for FP16 models using self-distillation.

\begin{table}[h]
  \caption{Recovery from pruning (FP16 models). PPL before and after 200 steps of LoRA training with self-distillation.}
  \label{tab:recovery}
  \centering
  \small
  \begin{tabular}{llccc}
    \toprule
    Model & Prune & No LoRA & With LoRA & Reduction \\
    \midrule
    Qwen-7B & 50\% & 920.6 & 24.3 & 97.4\% \\
    Qwen3-4B & 50\% & 148.2 & 43.2 & 70.9\% \\
    Nemotron & 30\% & 50.1 & 22.0 & 56.0\% \\
    Phi-2 & 50\% & 55.8 & 31.7 & 43.2\% \\
    \bottomrule
  \end{tabular}
\end{table}

Qwen2.5-7B with 50\% pruning goes from perplexity 920.56 (non-functional) to 24.27 after approximately two minutes on the L40S. Recovery scales with model size. LoRA fine-tuning on uncompressed models produces no measurable effect, confirming it functions purely as a recovery mechanism.

\paragraph{Distillation ablation.}
\label{sec:distill_ablation}
To determine whether the teacher signal contributes to recovery, we compared self-distillation against LoRA-only fine-tuning (cross-entropy loss, no teacher). Table~\ref{tab:distill_ablation} shows the results.

\begin{table}[h]
  \caption{Self-distillation vs.\ LoRA-only fine-tuning on pruned FP16 models, evaluated on full WikiText-2 (1{,}940 samples).}
  \label{tab:distill_ablation}
  \centering
  \small
  \begin{tabular}{llccc}
    \toprule
    Model & Prune & Pruned & Self-Distill & LoRA-Only \\
    \midrule
    Phi-2 & 50\% & 59.73 & 34.74 & 36.11 \\
    Qwen3-4B & 50\% & 146.19 & 46.45 & 39.71 \\
    Qwen-7B & 50\% & 1089.95 & 28.80 & 27.48 \\
    Nemotron & 30\% & 58.01 & 23.15 & 24.95 \\
    \bottomrule
  \end{tabular}
\end{table}

Both methods recover 40--97\% of pruning damage. The difference is small and inconsistent in direction: LoRA-only outperforms self-distillation for two models. The recovery is driven by LoRA adaptation, not the teacher signal, meaning the teacher model (which doubles memory during training) is not required.

\subsection{Compression Ordering}

For NF4 + pruning combinations, ordering determines viability. When pruning precedes quantization, NF4 models with pruning produce viable perplexity around 20--21. When quantization precedes pruning, the same combinations collapse. Among the 48 viable ordering configurations, ordering has no statistically significant effect on perplexity (one-way ANOVA, $F = 0.099$, $p = 0.96$).

\begin{table}[h]
  \caption{Perplexity by ordering (viable configs only). Missing entries indicate collapse.}
  \label{tab:ordering}
  \centering
  \small
  \begin{tabular}{lcc}
    \toprule
    Ordering & INT4 & INT8 \\
    \midrule
    $R \rightarrow P \rightarrow Q$ & 20.78 & 19.71 \\
    $P \rightarrow Q \rightarrow R$ & 20.67 & 19.73 \\
    $R \rightarrow Q \rightarrow P$ & -- & 21.45 \\
    $Q \rightarrow P \rightarrow R$ & -- & 20.75 \\
    \bottomrule
  \end{tabular}
\end{table}

\subsection{Benchmark Accuracy and Robustness}

We validated 30 compressed configurations on MMLU, HellaSwag, ARC-Challenge, and BoolQ. Every configuration with perplexity below 22 retained over 97\% of baseline average accuracy. INT8 quantization loses less than 0.5\% average accuracy; the recommended INT8 + 30\% pruning + recovery configuration loses under 1\% for three of four models and 0\% for Qwen2.5-7B. Multi-seed validation across three seeds showed standard deviations below 0.7 perplexity points, and the 20-sample evaluation subset correlates at $r = 0.985$ with the full WikiText-2 test set (1{,}940 samples). Full benchmark results, multi-seed tables, and the correlation plot are provided in the Appendix.

\subsection{On-Device CPU Deployment}
\label{sec:edge}

We deploy all four models on CPU via llama.cpp (see Section~\ref{sec:edge_setup} for methodology). Table~\ref{tab:edge} reports memory-tier feasibility.

\begin{table}[h]
  \caption{On-device CPU feasibility by memory tier (llama.cpp). RSS is peak resident set size; PPL is within-engine.}
  \label{tab:edge}
  \centering
  \small
  \begin{tabular}{llcccc}
    \toprule
    Tier & Models that fit & Example & RSS (MB) & tok/s & PPL \\
    \midrule
    2 GB & None & -- & -- & -- & -- \\
    4 GB & Phi-2 & Phi-2 q4 & 2848 & 17.5 & 11.01 \\
    8 GB & + Qwen3-4B, Nemotron & Nemotron q8 & 4356 & 9.8 & 9.25 \\
    12 GB & All four & Qwen2.5-7B q4 & 7545 & 7.3 & 6.91 \\
    \bottomrule
  \end{tabular}
\end{table}

Across all four models, 4-bit quantization runs 46--54\% faster than 8-bit with under 0.5 perplexity points of difference. Composing the study's GPU-side quantization with GGUF conversion is cheap (INT8-to-q8 adds under 0.1 PPL points; INT4-to-q4 adds 0.4--0.6), so the CPU results speak to the same compressed models evaluated on GPU. Unstructured pruning provides no CPU benefit: pruned variants run at the same throughput and RSS as non-pruned counterparts, because llama.cpp uses dense kernels and zeroed weights still occupy storage. Full per-configuration CPU measurements are in the Appendix.

\section{Discussion}
\label{sec:discussion}

\subsection{Practical Recommendations}

Table~\ref{tab:recommendations} summarizes the deployment-specific guidance that follows from the experimental findings.

\begin{table}[t]
  \caption{Deployment recommendations by target environment. All costs and benefits are measured in this study.}
  \label{tab:recommendations}
  \centering
  \small
  \begin{tabular}{>{\raggedright\arraybackslash}p{2.1cm}>{\raggedright\arraybackslash}p{2.4cm}>{\raggedright\arraybackslash}p{2.5cm}}
    \toprule
    Target & Config & Cost / benefit \\
    \midrule
    GPU server ($\geq$16\,GB) & INT8 only & $<$0.5\% acc.\ loss; 40--45\% memory \\
    \addlinespace
    GPU, sparse runtime & INT8 + 30\% prune + recovery & $<$1\% acc.\ loss; validate per arch. \\
    \addlinespace
    Mobile (4--6\,GB) & INT4, no pruning & 1--2.6\% acc.\ loss \\
    \addlinespace
    CPU, latency-bound & 4-bit (q4) & 46--54\% faster; $<$0.5 PPL vs.\ q8 \\
    \addlinespace
    Avoid & NF4 + pruning, quantize-first & collapse (16/16 configs) \\
    \bottomrule
  \end{tabular}
\end{table}

For GPU servers with 16GB+ memory, INT8 quantization alone is the simplest and most effective option: it provides 40--45\% memory reduction with no quality loss and matches the three-component recipe on accuracy (Section~\ref{sec:benchmark_val}) while avoiding the complexity of pruning and recovery. Unstructured pruning does not lower deployment cost on its own: in dense storage it reduces neither memory footprint, since zeroed weights still occupy the same number of bits, nor latency on dense kernels (Section~\ref{sec:edge}). Pruning pays off only with sparsity-aware execution: either a runtime that exploits unstructured sparsity directly (for example, DeepSparse on CPU), or re-pruning into a hardware-supported semi-structured pattern such as 2:4 sparsity on NVIDIA Ampere and later GPUs. Our unstructured magnitude pruning would need the latter conversion to benefit. For teams with such support, INT8 + 30\% pruning + recovery achieves under 1\% accuracy loss, and multi-seed validation confirms this recommendation is stable ($\sigma \leq 0.18$ across three seeds for this configuration; Section~\ref{sec:robustness}).

Mobile and embedded deployment with 4--6GB memory budgets narrows the options to INT4 without pruning (1--2.6\% accuracy loss). For CPU and latency-constrained targets, quantization level is the controlling variable: 4-bit quantization runs 46--54\% faster than 8-bit for under 0.5 perplexity points of difference (Section~\ref{sec:edge}, Figure~\ref{fig:edge_latency}), while pruning is counterproductive on dense CPU kernels.

Two cross-cutting findings inform these recommendations. First, LoRA-based fine-tuning without a teacher signal is sufficient for post-compression recovery (Section~\ref{sec:distill_ablation}, Table~\ref{tab:distill_ablation}), halving the memory required during the recovery training step relative to self-distillation. Second, the incremental benefit of pruning + recovery over INT8 quantization alone varies by model: Nemotron-Mini-4B showed no additional gain from this combination (see the technical supplement), so practitioners should validate the marginal benefit on their target architecture.

\subsection{Scope and Limitations}

Several design choices bound the generalizability of these findings. The on-device evaluation covers CPU inference via llama.cpp on a single machine class; mobile ARM and NPU hardware remain untested. The CPU study uses GGUF q4/q8 quantization rather than the bitsandbytes INT8/NF4 of the factorial study, and we report only within-engine comparisons as absolute perplexity values are not comparable across inference backends.

We evaluate unstructured pruning exclusively; structured alternatives (attention head pruning) proved inapplicable to three of four architectures (Section~\ref{sec:structured_pruning}). The full 120-configuration sweep uses a single seed, though multi-seed validation on 16 key configurations showed negligible variance (Section~\ref{sec:robustness}). Self-distillation for the 7B model requires INT8 quantization of the student to fit both models within 48GB VRAM, constraining the recovery options at that scale. Memory measurements during a live GPU session include runtime buffers and may not precisely reflect deployment requirements.

Finally, all benchmarks are English-only, and aggressive compression may disproportionately degrade performance on underrepresented inputs, an effect we do not measure and leave to future work.
\section{Conclusion}

Across 120 compression configurations on four architectures, three findings stand out. INT8 quantization halves memory with no accuracy cost and is the primary driver of memory reduction in our experiments. NF4 quantization combined with L1 magnitude pruning collapses models in all 16 tested combinations, stable across all seeds; one FP16+50\% configuration also collapsed. LoRA-based fine-tuning reduces degraded perplexity by up to 97\% in about two minutes of training, with recovery scaling by model size. An ablation comparing self-distillation against LoRA-only fine-tuning found no consistent benefit from the teacher signal, meaning the teacher model is not required for post-pruning recovery. Compression ordering does not matter among viable configurations, but determines viability for NF4+pruning combinations: pruning must precede quantization. Perplexity on a 20-sample subset correlates at $r = 0.985$ with full-corpus evaluation, validating it as a fast screening metric. On-device CPU deployment confirms these patterns transfer off the GPU: 4-bit quantization runs 46--54\% faster than 8-bit for under half a perplexity point of difference, while unstructured pruning provides no CPU latency or memory benefit. Future work should extend this analysis to newer architectures (Llama-3, Mistral, Gemma-2), compare across quantization backends (GPTQ, AWQ) and pruning methods (Wanda, SparseGPT), and extend on-device measurement to mobile ARM and NPU hardware.

\section*{Customer Impact Statement}

Teams deploying small language models (2B--7B parameters) on GPU servers, edge devices, and CPU endpoints face a combinatorial search across quantization, pruning, and distillation with no systematic guidance on how these techniques interact. We replace that trial-and-error with a tested framework spanning 120 compression configurations across four architectures. INT8 quantization alone provides 40--45\% memory reduction with under 0.5\% accuracy loss. Adding 30\% pruning with LoRA-based fine-tuning maintains under 1\% accuracy loss; the LoRA recovery step does not require a teacher model, halving memory during recovery training. On-device CPU measurements map four memory tiers (2, 4, 8, 12 GB) to model feasibility and show 4-bit quantization runs 46--54\% faster than 8-bit. The failure modes are equally actionable: NF4 quantization combined with L1 pruning collapses models in all 16 tested combinations, producing fluent-looking but incoherent output. Scope limitations: all benchmarks are English-only; architecture-specific behavior varies (Nemotron-Mini-4B saw no benefit from the recommended recipe); on-device measurements cover CPU on a single machine class; aggressive compression may amplify biases on underrepresented inputs, which we did not measure.


\bibliographystyle{plainnat}
\bibliography{references}

\appendix
\section{Benchmark Accuracy on Compressed Models}
\label{sec:benchmark_val}

Table~\ref{tab:benchmark} presents per-model benchmark accuracy for a representative subset of 30 compressed configurations. Bold rows indicate the recommended compression configuration per model (INT8 + 30\% pruning + recovery). $\Delta$Acc shows the change in average accuracy relative to the uncompressed FP16 baseline.

\begin{table}[t]
  \caption{Benchmark accuracy on compressed models. Avg Acc is the mean of MMLU, HellaSwag, ARC, and BoolQ. $\Delta$ shows change from baseline. Bold indicates best compressed config per model.}
  \label{tab:benchmark}
  \centering
  \footnotesize
  \begin{tabular}{llllccccccr}
    \toprule
    Model & Quant & Prune & Recov & PPL & MMLU & HSwag & ARC & BoolQ & Avg Acc & $\Delta$Acc \\
    \midrule
    \multicolumn{11}{l}{\textit{Phi-2 (2.7B)}} \\
    & none & none & no & 20.67 & 0.545 & 0.559 & 0.530 & 0.834 & 0.617 & -- \\
    & none & none & yes & 20.91 & 0.543 & 0.556 & 0.534 & 0.836 & 0.617 & 0.0\% \\
    & INT8 & none & no & 21.25 & 0.543 & 0.555 & 0.531 & 0.829 & 0.615 & $-$0.3\% \\
    & INT8 & none & yes & 20.99 & 0.542 & 0.554 & 0.522 & 0.837 & 0.614 & $-$0.5\% \\
    & \textbf{INT8} & \textbf{30\%} & \textbf{yes} & \textbf{21.14} & \textbf{0.531} & \textbf{0.551} & \textbf{0.522} & \textbf{0.834} & \textbf{0.610} & $\mathbf{-1.1\%}$ \\
    & INT4 & none & no & 21.99 & 0.532 & 0.547 & 0.513 & 0.812 & 0.601 & $-$2.6\% \\
    & INT4 & none & yes & 21.86 & 0.532 & 0.549 & 0.514 & 0.817 & 0.603 & $-$2.3\% \\
    & none & 30\% & yes & 22.82 & 0.519 & 0.541 & 0.506 & 0.818 & 0.596 & $-$3.4\% \\
    \midrule
    \multicolumn{11}{l}{\textit{Qwen3-4B}} \\
    & none & none & no & 26.72 & 0.684 & 0.523 & 0.508 & 0.851 & 0.642 & -- \\
    & none & none & yes & 26.53 & 0.684 & 0.523 & 0.509 & 0.851 & 0.642 & 0.0\% \\
    & INT8 & none & no & 27.25 & 0.680 & 0.522 & 0.511 & 0.851 & 0.641 & $-$0.2\% \\
    & \textbf{INT8} & \textbf{30\%} & \textbf{yes} & \textbf{27.31} & \textbf{0.681} & \textbf{0.518} & \textbf{0.499} & \textbf{0.851} & \textbf{0.637} & $\mathbf{-0.8\%}$ \\
    & INT4 & none & no & 28.82 & 0.670 & 0.511 & 0.493 & 0.846 & 0.630 & $-$1.9\% \\
    & none & 30\% & yes & 28.09 & 0.663 & 0.499 & 0.479 & 0.845 & 0.622 & $-$3.1\% \\
    \midrule
    \multicolumn{11}{l}{\textit{Qwen2.5-7B-Instruct}} \\
    & none & none & no & 11.82 & 0.718 & 0.621 & 0.527 & 0.864 & 0.683 & -- \\
    & INT8 & none & no & 11.82 & 0.720 & 0.622 & 0.528 & 0.862 & 0.683 & 0.0\% \\
    & \textbf{INT8} & \textbf{30\%} & \textbf{yes} & \textbf{12.29} & \textbf{0.717} & \textbf{0.619} & \textbf{0.528} & \textbf{0.866} & \textbf{0.683} & $\mathbf{0.0\%}$ \\
    & INT4 & none & no & 12.74 & 0.711 & 0.613 & 0.511 & 0.869 & 0.676 & $-$1.0\% \\
    & INT4 & none & yes & 12.69 & 0.710 & 0.613 & 0.513 & 0.869 & 0.676 & $-$1.0\% \\
    \midrule
    \multicolumn{11}{l}{\textit{Nemotron-Mini-4B}} \\
    & none & none & no & 18.53 & 0.569 & 0.559 & 0.506 & 0.826 & 0.615 & -- \\
    & none & none & yes & 18.51 & 0.568 & 0.559 & 0.506 & 0.825 & 0.615 & 0.0\% \\
    & INT8 & none & no & 18.51 & 0.567 & 0.558 & 0.503 & 0.826 & 0.614 & $-$0.2\% \\
    & \textbf{INT8} & \textbf{30\%} & \textbf{yes} & \textbf{18.89} & \textbf{0.567} & \textbf{0.558} & \textbf{0.509} & \textbf{0.836} & \textbf{0.618} & $\mathbf{+0.5\%}$ \\
    & INT4 & none & no & 19.59 & 0.558 & 0.549 & 0.500 & 0.823 & 0.608 & $-$1.1\% \\
    & none & 30\% & yes & 22.02 & 0.529 & 0.527 & 0.473 & 0.820 & 0.587 & $-$4.5\% \\
    \bottomrule
  \end{tabular}
\end{table}

\section{Multi-Seed Validation}
\label{sec:robustness}

We repeated 16 key configurations (4 per model) across three random seeds (42, 123, 456) on the full WikiText-2 test set. Configurations without stochastic components produced identical perplexity. Self-distillation configurations showed $\sigma < 0.7$ PPL points (Table~\ref{tab:multiseed}).

\begin{table}[ht]
  \caption{Multi-seed validation (3 seeds, full WikiText-2).}
  \label{tab:multiseed}
  \centering
  \small
  \begin{tabular}{llcc}
    \toprule
    Model & Config & PPL $\mu$ & $\sigma$ \\
    \midrule
    Phi-2 & Baseline & 22.28 & 0.00 \\
    Phi-2 & INT8+30\%+R & 22.86 & 0.07 \\
    Phi-2 & INT4 & 23.62 & 0.00 \\
    Phi-2 & FP16+50\%+R & 34.79 & 0.28 \\
    Qwen3-4B & Baseline & 26.89 & 0.00 \\
    Qwen3-4B & INT8+30\%+R & 27.35 & 0.18 \\
    Qwen3-4B & INT4 & 28.75 & 0.00 \\
    Qwen3-4B & FP16+50\%+R & 47.01 & 0.66 \\
    Qwen-7B & Baseline & 14.54 & 0.00 \\
    Qwen-7B & INT8+30\%+R & 15.08 & 0.02 \\
    Qwen-7B & INT4 & 15.50 & 0.00 \\
    Qwen-7B & FP16+50\%+R & 28.84 & 0.23 \\
    Nemotron & Baseline & 20.27 & 0.00 \\
    Nemotron & INT8+30\%+R & 20.34 & 0.04 \\
    Nemotron & INT4 & 21.37 & 0.00 \\
    Nemotron & FP16+50\%+R & 41.87 & 0.16 \\
    \bottomrule
  \end{tabular}
\end{table}

\section{Full-Corpus Perplexity Validation}

\begin{figure}[h]
  \centering
  \includegraphics[width=\columnwidth]{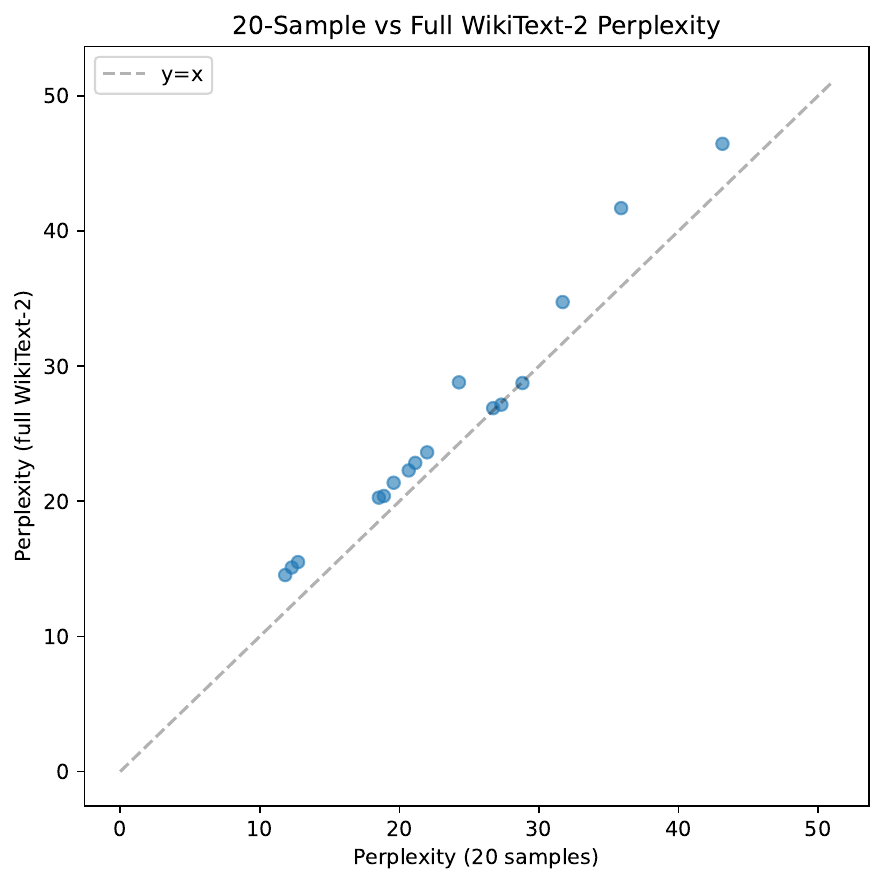}
  \caption{20-sample vs full WikiText-2 perplexity across 16 configurations. $r = 0.985$.}
  \label{fig:correlation}
\end{figure}

The Pearson correlation between 20-sample and full-corpus (1{,}940 sample) perplexity is $r = 0.985$, with mean absolute difference 2.21 points and maximum difference 5.80 points (Figure~\ref{fig:correlation}). Configuration rankings are fully preserved.

\section{Structured Pruning}
\label{sec:structured_pruning}

Attention head pruning was evaluated as a structured alternative. Only Nemotron-Mini-4B exposed head-level parameters; structured pruning at 30\% produced perplexity 4{,}775 (vs 52.69 unstructured) and collapsed at 50\%. The other architectures use fused attention projections incompatible with external head-level pruning. This motivated unstructured pruning throughout the study.

\section{Statistical Analysis}

We fit an ANOVA on 55 viable configurations with model as a blocking factor. No factor reaches significance: quantization ($\eta^2 = 0.04$, $p = 0.33$), pruning ($\eta^2 = 0.07$, $p = 0.14$), recovery ($\eta^2 = 0.03$, $p = 0.21$), model ($\eta^2 = 0.03$, $p = 0.63$). The residual (82\% of variance) reflects the dominant $Q \times P$ interaction: INT4+pruning causes collapse while other combinations cluster in 20--30 perplexity, a binary boundary rather than gradual degradation. Per-model, pruning is the largest factor for three of four models ($\eta^2$ up to 0.36) and quantization for Nemotron ($\eta^2 = 0.28$); recovery is smallest for all ($\eta^2 \leq 0.08$).

\section{On-Device CPU Benchmark (Full Results)}

\begin{figure}[h]
  \centering
  \includegraphics[width=\columnwidth]{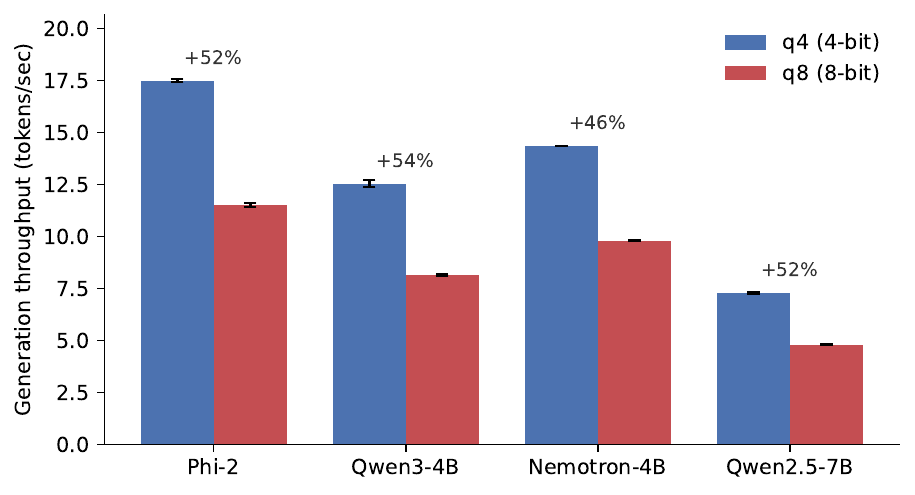}
  \caption{Measured CPU generation throughput (llama.cpp), 4-bit versus 8-bit, with standard-deviation error bars over five runs. 4-bit runs 46--54\% faster across all four models, with negligible variance.}
  \label{fig:edge_latency}
\end{figure}

Table~\ref{tab:edge_speed} summarizes the 4-bit vs 8-bit tradeoff, and Table~\ref{tab:edge_full} lists all measured CPU configurations. Figure~\ref{fig:edge_latency} shows the throughput comparison.

\begin{table}[ht]
  \caption{4-bit vs 8-bit on CPU (llama.cpp). Speedup is q4 over q8.}
  \label{tab:edge_speed}
  \centering
  \small
  \begin{tabular}{lccccc}
    \toprule
    Model & q4 tok/s & q8 tok/s & Speedup & q4 PPL & q8 PPL \\
    \midrule
    Phi-2 & 17.5 & 11.5 & +52\% & 11.01 & 10.74 \\
    Qwen3-4B & 12.5 & 8.1 & +54\% & 14.57 & 14.15 \\
    Nemotron-4B & 14.4 & 9.8 & +46\% & 9.50 & 9.25 \\
    Qwen2.5-7B & 7.3 & 4.8 & +52\% & 6.91 & 7.31 \\
    \bottomrule
  \end{tabular}
\end{table}

\begin{table}[ht]
  \caption{Full on-device CPU measurements (llama.cpp).}
  \label{tab:edge_full}
  \centering
  \small
  \begin{tabular}{lccccc}
    \toprule
    Config & RSS (MB) & tok/s & $\sigma$ & PPL & Min fit \\
    \midrule
    \multicolumn{6}{l}{\textit{Phi-2 (2.7B)}} \\
    phi2-baseline-q4 & 2848 & 17.50 & 0.09 & 11.01 & 4 GB \\
    phi2-baseline-q8 & 2957 & 11.53 & 0.10 & 10.74 & 4 GB \\
    phi2-int4-q4 & 2848 & 17.93 & 0.07 & 11.64 & 4 GB \\
    phi2-int4-d-q4 & 2848 & 17.94 & 0.13 & 11.65 & 4 GB \\
    phi2-int8-q8 & 2957 & 11.48 & 0.09 & 10.85 & 4 GB \\
    phi2-int8-p30-d-q8 & 2957 & 11.69 & 0.03 & 13.04 & 4 GB \\
    \midrule
    \multicolumn{6}{l}{\textit{Qwen3-4B}} \\
    qwen3-4b-baseline-q4 & 4183 & 12.53 & 0.17 & 14.57 & 8 GB \\
    qwen3-4b-baseline-q8 & 4213 & 8.14 & 0.03 & 14.15 & 8 GB \\
    qwen3-4b-int4-q4 & 4183 & 12.74 & 0.03 & 15.19 & 8 GB \\
    qwen3-4b-int8-q8 & 4213 & 8.11 & 0.07 & 14.13 & 8 GB \\
    qwen3-4b-int8-p30-d-q8 & 4213 & 8.15 & 0.07 & 11.34 & 8 GB \\
    \midrule
    \multicolumn{6}{l}{\textit{Nemotron-Mini-4B}} \\
    nemotron-4b-baseline-q4 & 3794 & 14.35 & 0.02 & 9.50 & 8 GB \\
    nemotron-4b-baseline-q8 & 4356 & 9.80 & 0.04 & 9.25 & 8 GB \\
    nemotron-4b-int8-q8 & 4356 & 9.77 & 0.00 & 9.25 & 8 GB \\
    \midrule
    \multicolumn{6}{l}{\textit{Qwen2.5-7B-Instruct}} \\
    qwen25-7b-baseline-q4 & 7545 & 7.29 & 0.04 & 6.91 & 12 GB \\
    qwen25-7b-baseline-q8 & 7862 & 4.79 & 0.03 & 7.31 & 12 GB \\
    qwen25-7b-int4-q4 & 7546 & 7.63 & 0.04 & 7.30 & 12 GB \\
    qwen25-7b-int8-q8 & 7861 & 4.75 & 0.03 & 7.30 & 12 GB \\
    qwen25-7b-int8-p30-d-q8 & 7862 & 4.69 & 0.01 & 9.55 & 12 GB \\
    \bottomrule
  \end{tabular}
\end{table}

\end{document}